\documentclass[conference]{IEEEtran}
\IEEEoverridecommandlockouts

\newif\ifieee
\ieeefalse

\usepackage{cite}
\usepackage{amsmath,amssymb,amsfonts}
\usepackage{algorithmic}
\usepackage{graphicx}
\usepackage{textcomp}
\usepackage[dvipsnames,table,xcdraw]{xcolor}
\def\BibTeX{{\rm B\kern-.05em{\sc i\kern-.025em b}\kern-.08em
    T\kern-.1667em\lower.7ex\hbox{E}\kern-.125emX}}

\usepackage[caption=false,farskip=0pt]{subfig}
\usepackage{booktabs}
\usepackage{multirow}
\usepackage[hyphens]{url}
\ifieee
\else
\usepackage[colorlinks=true,allcolors=black]{hyperref}
\fi
\usepackage[acronym]{glossaries}
\usepackage{xspace}
\usepackage[T1]{fontenc}
\usepackage[capitalise]{cleveref} %
\usepackage{transparent}
\usepackage{moresize}
\usepackage[normalem]{ulem}
\usepackage{soul} %
\setuldepth{Berlin}
\usepackage{balance}
\usepackage{CJKutf8}
\usepackage{comment}

\newif\ifnocomments
\nocommentstrue

\usepackage{tikz}

\newcommand\copyrighttext{%
  \scriptsize Accepted at SIBGRAPI 2022. The final published version is available on IEEE Xplore (DOI: \href{https://doi.org/10.1109/SIBGRAPI55357.2022.9991768}{\textcolor{blue}{10.1109/SIBGRAPI55357.2022.9991768}}).}
\newcommand\copyrightnotice{%
\begin{tikzpicture}[remember picture,overlay]
\node[anchor=south,yshift=30pt,xshift=0pt] at (current page.south) {\fbox{\transparent{0.85}\parbox{\dimexpr0.76\textwidth-\fboxsep-\fboxrule\relax}{\copyrighttext}}};
\end{tikzpicture}%
}

\newif\ifextended
\extendedfalse

\newif\ifalt
\altfalse

\ifnocomments
\newcommand*{\RL}[2][]{}
\newcommand*{\DM}[2][]{}
\newcommand*{\VE}[2][]{}
\newcommand*{\todo}[2][]{}

\else

\newcommand*{\RL}[2][]{\textcolor{violet}{[\textbf{\ifthenelse{\equal{#1}{}}{RL}{RL(#1)}}: #2]}}
\newcommand\DM[1]{{\textcolor{blue}{[\textbf{DM: #1}]}}}
\newcommand\VE[1]{{\textcolor{ForestGreen}{[\textbf{VE: #1}]}}}
\newcommand*{\todo}[2][]{\textcolor{red}{[\textbf{\ifthenelse{\equal{#1}{}}{TODO}{TODO(#1)}}: #2]}}

\fi

\newcommand*{\tofinish}[2][]{\textcolor{BurntOrange}{[\textbf{\ifthenelse{\equal{#1}{}}{TODO}{TODO(#1)}}: #2]}}

\newcommand{\highlight}[3][black]{{\transparent{0.6}\fboxsep0.5pt\colorbox{#2}{\color{#1} #3}}}

\newcounter{fncounter}
\newcommand\customfootnote[1]{\stepcounter{fncounter}\footnote{\hspace{0.2mm}#1}}

\newcounter{fnAnnotations}
\newacronymstyle{long-short-br}
{%
  \GlsUseAcrEntryDispStyle{long-short}%
}%
{%
  \GlsUseAcrStyleDefs{long-short}%
}
\setacronymstyle{long-short-br}

\newif\iffinal
\finaltrue
\newcommand{\cmtid}{85}

\iffinal
\else
\usepackage[switch]{lineno}
\fi

\begin{document}

\newcommand{\papertitle}{\huge A First Look at Dataset Bias in License Plate Recognition}

\iffinal

\title{\papertitle}

\author{Rayson Laroca\IEEEauthorrefmark{1}, Marcelo Santos\IEEEauthorrefmark{1}, Valter Estevam\IEEEauthorrefmark{1}\IEEEauthorrefmark{2}, Eduardo Luz\IEEEauthorrefmark{3}, and David Menotti\IEEEauthorrefmark{1}\\
\IEEEauthorrefmark{1}\hspace{0.15mm}Department of Informatics, Federal University of Paran\'a, Curitiba, Brazil\\
\IEEEauthorrefmark{2}\hspace{0.15mm}Federal Institute of Paran\'a, Irati, Brazil\\
\IEEEauthorrefmark{3}\hspace{0.15mm}Department of Computer Science, Federal University of Ouro Preto,
Ouro Preto, Brazil\\
\resizebox{0.95\linewidth}{!}{
\hspace{0.3mm}\IEEEauthorrefmark{1}{\hspace{-0.15mm}\tt\small \{rblsantos,msantos,vlejunior,menotti\}@inf.ufpr.br} \quad \IEEEauthorrefmark{2}{\hspace{0.15mm}\tt\small valter.junior@ifpr.edu.br} \quad \IEEEauthorrefmark{3}{\hspace{0.15mm}\tt\small eduluz@ufop.edu.br}
}
}

\else
    \title{\papertitle}

    \author{SIBGRAPI paper ID: \cmtid }
\fi

\maketitle

\ifieee
{\let\thefootnote\relax\footnote{\\978-1-6654-5385-1/22/\$31.00 \textcopyright2022 IEEE}}
\else
\copyrightnotice
\transparent{1}
\fi

\newacronym{alpr}{ALPR}{Automatic License Plate Recognition}
\newacronym{auc}{AUC}{Area Under the Curve}
\newacronym{cnn}{CNN}{Convolutional Neural Network}
\newacronym{denatran}{DENATRAN}{National Traffic Department of Brazil}
\newacronym{gan}{GAN}{Generative Adversarial Network}
\newacronym{lp}{LP}{license plate}
\newacronym{lodo}{LODO}{leave-one-dataset-out}
\newacronym{lpd}{LPD}{License Plate Detection}
\newacronym{lpr}{LPR}{License Plate Recognition}
\newacronym{lstm}{LSTM}{Long Short-Term Memory}
\newacronym{iou}{IoU}{Intersection over Union}
\newacronym{it}{IT}{information technology}
\newacronym{mse}{MSE}{Mean Squared Error}
\newacronym{ocr}{OCR}{Optical Character Recognition}
\newacronym{relu}{ReLU}{Rectified Linear Unit}
\newacronym{roc}{ROC}{Receiver Operating Characteristic}
\newacronym{rodosol}{RodoSol}{\textit{Rodovia do Sol}}
\newacronym{sgd}{SGD}{Stochastic Gradient Descent}
\newacronym{svm}{SVM}{Support Vector Machine}

\newcommand{\ccpd}{CCPD\xspace}
\newcommand{\chineselp}{ChineseLP\xspace}
\newcommand{\clpd}{CLPD\xspace}
\newcommand{\pku}{PKU\xspace}
\newcommand{\platesmaniaietcn}{PlatesMania-CN\xspace}
\newcommand{\rodosolalpr}{RodoSol-ALPR\xspace}
\newcommand{\ssigsegplate}{SSIG-SegPlate\xspace}
\newcommand{\ufop}{UFOP\xspace}
\newcommand{\ufpralpr}{UFPR-ALPR\xspace}

\newcommand{\model}{DC-NET\xspace}
\newcommand{\cdcc}{CDCC-NET\xspace}

\newcommand{\holistic}{Holistic-CNN\xspace}
\newcommand{\crnn}{CRNN\xspace}
\newcommand{\trba}{TRBA\xspace}
\newcommand{\gradcam}{Grad-CAM\xspace}
\newcommand{\bracc}{95.2\xspace}
\newcommand{\cnacc}{95.9\xspace}

\newcommand{\bothAccMoreAbs}{95}
\newcommand{\bothAccMore}{\bothAccMoreAbs\xspace}

\newcommand{\accRodosol}{99.7\xspace}
\newcommand{\accUfpr}{85.5\xspace}

\newcommand{\gameavgconf}{95.9\xspace}
\newcommand{\gamestdevconf}{9.3\xspace}

\iffinal
\newcommand{\supplementarySplits}{\url{https://raysonlaroca.github.io/supp/sibgrapi2022/splits.zip}}
\newcommand{\supplementaryAnnotations}{\url{https://raysonlaroca.github.io/supp/sibgrapi2022/annotations.zip}}

\else

\newcommand{\supplementarySplits}{[hidden for review].}

\fi

\ifieee
\vspace{-4mm}
\else
\vspace{-4mm} %
\fi
\begin{abstract}
Public datasets have played a key role in advancing the state of the art in \gls*{lpr}.
Although dataset bias has been recognized as a severe problem in the computer vision community, it has been largely overlooked in the \gls*{lpr} literature.
\gls*{lpr} models are usually trained and evaluated separately on each dataset.
In this scenario, they have often proven robust in the dataset they were trained in but showed limited performance in unseen ones.
Therefore, this work investigates the dataset bias problem in the \gls*{lpr} context.
We performed experiments on eight datasets, four collected in Brazil and four in mainland China, and observed that each dataset has a unique, identifiable ``signature'' since a lightweight classification model predicts the source dataset of a \gls*{lp} image with more than \bothAccMoreAbs\% accuracy.
In our discussion, we draw attention to the fact that most \gls*{lpr} models are probably exploiting such signatures to improve the results achieved in each dataset at the cost of losing generalization capability.
These results emphasize the importance of evaluating \gls*{lpr} models in cross-dataset setups, as they provide a better indication of generalization (hence real-world performance) than within-dataset~ones.

\end{abstract}

\section{Introduction}
\label{sec:introduction}

\glsresetall

Is it possible to predict the dataset from which a \gls*{lp} image is coming?
Initially, one may think that this task is fairly trivial since --~in principle~-- images from distinct datasets are collected in different regions, with different hardware, for different purposes, etc.
On second thought, one may realize that it depends on the datasets we are comparing.

Suppose there are two datasets, one composed exclusively of images of American \glspl*{lp} and the other of images of European \glspl*{lp}.
In that case, it should indeed be relatively straightforward to distinguish which dataset each \gls*{lp} image belongs to due to the many characteristics \glspl*{lp} from the same region/layout share in common, e.g., the aspect ratio, colors, symbols, the position of the characters, the number of characters, among others (this is well illustrated in~\cite{laroca2021efficient}).
Nevertheless, beyond the \gls*{lp} layout, are there unique signatures (\emph{bias}) in each dataset that would enable identifying the source of an \gls*{lp}~image?

The presence of unique signatures in public datasets was first revealed by Torralba et al.~\cite{torralba2011unbiased}.
They investigated the then-popular object recognition datasets (PASCAL'07, ImageNet, among others) using the \emph{Name That Dataset!} experiment in which a \gls*{svm} classifier was trained to distinguish images from 12 datasets.
If \emph{dataset bias} did not exist, no classifier would be able to perform this task at levels considerably different from chance.
However, their classifier reached an accuracy of $39$\%, which is significantly better than chance ($1/12$ = $8$\%).
This result becomes even more surprising when taking into account that those datasets were created with the expressed goal of being as varied and rich as possible, aiming to sample the visual world ``in the wild''~\cite{torralba2011unbiased}.

Dataset bias has been consistently recognized as a severe problem in the computer vision community~\cite{tommasi2017deeper,ashraf2018learning,wachinger2021detect,jaipuria2022deeppic}, as models are inadvertently learning idiosyncrasies of each dataset along with knowledge fundamental to the task under study.
Nevertheless, to the best of our knowledge, it has remained largely unnoticed in the \gls*{lpr} literature.

\begin{figure}[t]
	\centering
	\captionsetup[subfigure]{labelformat=empty,captionskip=-28.5pt,margin=3.6pt,justification=raggedleft,singlelinecheck=false}

        \vspace{0.75mm}

	\resizebox{0.975\linewidth}{!}{
		\subfloat[\highlight{white}{\ssmall (a)}]{
			\includegraphics[width=0.32\linewidth]{    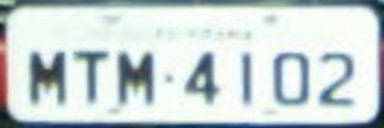} 
		} 
		\hspace{-2.5mm}
		\subfloat[\highlight{white}{\ssmall (b)}]{
			\includegraphics[width=0.32\linewidth]{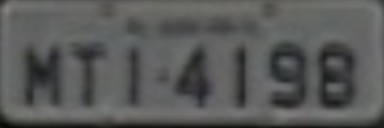}
		}
		\hspace{-2.5mm}
		\subfloat[\highlight{white}{\ssmall (c)}]{
    \includegraphics[width=0.32\linewidth]{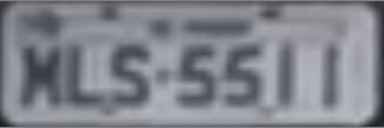}
			} \,

	}

	\vspace{0.9mm}

	\resizebox{0.975\linewidth}{!}{
		\subfloat[\highlight{white}{\ssmall (d)}]{
			\includegraphics[width=0.32\linewidth]{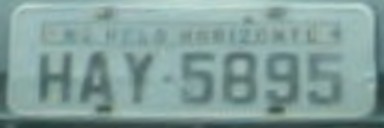}
		}
		\hspace{-2.5mm}
		\subfloat[\highlight{white}{\ssmall (e)}]{
			\includegraphics[width=0.32\linewidth]{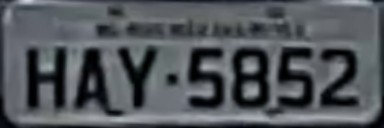}
		}
		\hspace{-2.5mm}
		\subfloat[\highlight{white}{\ssmall (f)}]{
			\includegraphics[width=0.32\linewidth]{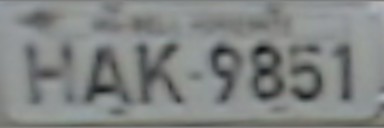}
			} \,
	}

	\vspace{0.9mm}

	\resizebox{0.975\linewidth}{!}{
		\subfloat[\highlight{white}{\ssmall (g)}]{
			\includegraphics[width=0.32\linewidth]{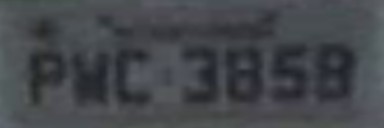}
		}
		\hspace{-2.5mm}
		\subfloat[\highlight{white}{\ssmall (h)}]{
			\includegraphics[width=0.32\linewidth]{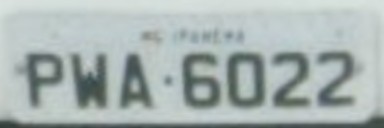}
		}
		\hspace{-2.5mm}
		\subfloat[\highlight{white}{\ssmall (i)}]{
			\includegraphics[width=0.32\linewidth]{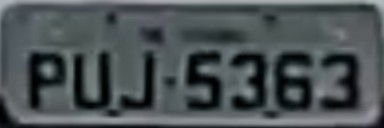}
			} \,
	}

	\vspace{0.9mm}

	\resizebox{0.975\linewidth}{!}{
		\subfloat[\highlight{white}{\ssmall (j)}]{
			\includegraphics[width=0.32\linewidth]{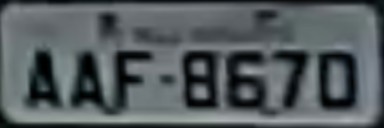}
		}
		\hspace{-2.5mm}
		\subfloat[\highlight{white}{\ssmall (k)}]{
			\includegraphics[width=0.32\linewidth]{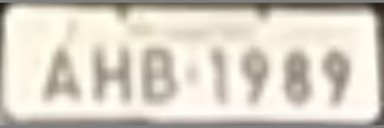}
		}
		\hspace{-2.5mm}
		\subfloat[\highlight{white}{\ssmall (l)}]{
        \includegraphics[width=0.32\linewidth]{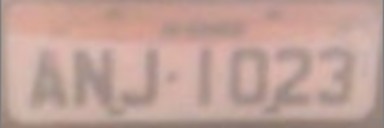}
		    } \, 
	}

	\vspace{0.9mm}

	\resizebox{0.975\linewidth}{!}{
		\subfloat[\highlight{white}{\ssmall (m)}]{
			\includegraphics[width=0.32\linewidth]{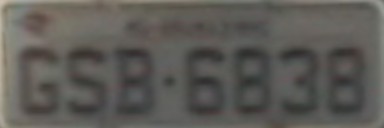}
		}
		\hspace{-2.5mm}
		\subfloat[\highlight{white}{\ssmall (n)}]{
			\includegraphics[width=0.32\linewidth]{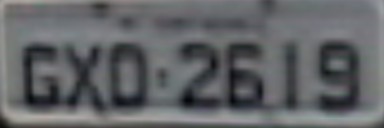}
		}
		\hspace{-2.5mm}
		\subfloat[\highlight{white}{\ssmall (o)}]{
			\includegraphics[width=0.32\linewidth]{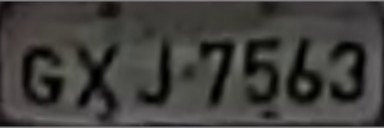}
			} \,
	}
	    
    \vspace{0.3mm}
	    
	\resizebox{0.975\linewidth}{!}{
		\begin{tabular}{rr}
			\rodosolalpr (ES)~\cite{laroca2022cross}: \_\_\_~, \_\_\_~, \_\_\_~, \_\_\_~ & \ssigsegplate (MG)~\cite{goncalves2016benchmark}: \_\_\_~, \_\_\_~, \_\_\_~, \_\_\_\;\;\,
			\\[1pt]
			\ufop (MG)~\cite{mendesjunior2011towards}: \_\_\_~, \_\_\_~, \_\_\_~, \_\_\_~        & \ufpralpr (PR)~\cite{laroca2018robust}: \_\_\_~, \_\_\_~, \_\_\_\phantom{~, \_\_\_\;\;\,}          
		\end{tabular}%
	}
	
	\vspace{-0.75mm}
	
    \caption[Caption for LOF]{Can you name the dataset to which each of the above images belongs?
    (you can try grouping the images into four distinct groups if you are unfamiliar with the corresponding datasets).
    See footnote\footnotemark[\thefncounter] for the answer key.
    This task is somewhat challenging for humans, as \gls*{lp} images from distinct datasets have similar characteristics.
    However, a shallow CNN ($3$ conv. layers) predicts the correct dataset in more than $\bothAccMore$\% of cases (chance is $1/4$ = $25$\%).
    All images above were classified correctly, with a mean confidence value of~$\gameavgconf$\%.
    }
	\newcommand\fngame{\thefncounter}
	\newcounter{fngame}
	\setcounter{fngame}{\thefncounter}
	\label{fig:name-that-dataset}
\end{figure}
\footnotetext[\thefngame]{\footnotesize Answer key: \rodosolalpr $\rightarrow$ (a),(d),(h),(l); \ssigsegplate $\rightarrow$ (e),(i),(j),(o); \ufop $\rightarrow$ (b),(f),(m),(n); and \ufpralpr $\rightarrow$ (c),(g),(k).}
\setcounter{footnote}{1}

Considering the above discussion, in this work we revisit the experiments conducted by Torralba et al.~\cite{torralba2011unbiased}, adapting them to the \gls*{lpr} context (see \cref{fig:name-that-dataset}, where we recreate the \emph{Name That Dataset!} game~\cite{torralba2011unbiased} with Brazilian \glspl*{lp}\customfootnote{To maintain consistency with previous work~\cite{silva2017realtime,laroca2018robust,oliveira2021vehicle},
we refer to ``Brazilian'' as the layout used in Brazil \emph{before} the adoption of the Mercosur~layout.}).
Our experiments, performed on public datasets acquired in Brazil and mainland China, demonstrate that a lightweight \gls*{cnn} can identify the source dataset of an LP image with more than $\bothAccMore$\% accuracy, which is much higher than expected from chance or human perceptual similarity judgments.
Intriguingly, our experiments also show no signs of saturation as more training data is added, i.e., the classification accuracy could be even higher if there were more training~data.

The severity of the dataset bias problem in \gls*{lpr} boils down to the following.
\gls*{lpr} datasets are usually very unbalanced in terms of character classes due to \gls*{lp} assignment  policies~\cite{anagnostopoulos2006license,zhang2021robust_attentional}.
In a dataset collected in Brazil, for instance, one letter may appear much more frequently than others according to the state in which most vehicles were registered~\cite{goncalves2018realtime,laroca2018robust} (e.g., the \ssigsegplate dataset~\cite{goncalves2016benchmark} has $746$ instances of the letter~`O' but only $135$ instances of the letter~`Q').
The same is true for vehicles registered in different cities within a province in mainland China~\cite{zhang2021robust_attentional,wang2022rethinking}.
Considering that \gls*{lpr} models are generally trained and evaluated on images from the same dataset (see \cref{sec:related_work}), such bias can skew the predictions towards the prominent character classes, resulting in poor generalization performance in the real world~\cite{yang2018chinese,zhang2020robust_license,laroca2022cross}.

The aim of this paper is two-fold.
First, to situate the dataset bias problem in the \gls*{lpr} context and thus raise awareness in the community regarding the possible impacts of such bias as this issue is not getting the attention it deserves.
Second, to discuss some subtle ways bias may have crept into the chosen datasets to outline directions for future~research.

\ifextended
The rest of the paper is organized as follows.
We review related work in \cref{sec:related_work}.
In \cref{sec:experiments}, we describe the experiments performed and also present the results obtained.
In \cref{sec:discussion}, we shed light on the impacts of dataset bias on the cross-dataset generalization of \gls*{lpr} models and also provide some insights into possible causes.
Finally, we conclude the paper and outline future research directions in \cref{sec:conclusions}.

\else
\fi
\section{Related Work}
\label{sec:related_work}

\gls*{alpr} systems exploit image processing and pattern recognition techniques to find and recognize \glspl*{lp} in images or videos~\cite{silva2022flexible,wang2022rethinking}.
Some practical applications for an \gls*{alpr} system are automatic toll collection, road traffic monitoring, traffic law enforcement, and vehicle access control in restricted areas~\cite{du2013automatic,weihong2020research}.

Deep learning-based \gls*{alpr} systems typically include two stages: \gls*{lpd} and \gls*{lpr}~\cite{goncalves2018realtime,xu2018towards}.
The former refers to locating the \gls*{lp} regions in the input image, while the latter refers to recognizing the characters on each \gls*{lp}.
Recent works have focused on the \gls*{lpr} stage~\cite{zeni2020weakly,liu2021fast,zhang2021robust_attentional}, as general-purpose object detectors have achieved impressive results in the \gls*{lpd} stage for some years now~\cite{hsu2017robust,laroca2018robust}.

The standard method of evaluating an \gls*{lpr} method's performance is to use multiple publicly available datasets, such as \ssigsegplate~\cite{goncalves2016benchmark} and \ccpd~\cite{xu2018towards}, which are split into disjoint training and test sets.
Such an assessment is typically done independently for each dataset~\cite{laroca2018robust,weihong2020research,zhang2021efficient}.
As models based on deep learning can take significant time to be trained, some authors have adopted a slightly different protocol where the proposed networks are trained once on the union of the training images from the chosen datasets and evaluated individually on the respective test sets~\cite{selmi2020delpdar,laroca2022cross,silva2022flexible}.
Although the images for training and testing belong to disjoint subsets, these protocols do not make it clear whether the evaluated models have good generalization ability, i.e., whether they perform well on images from other scenarios/datasets, mainly due to domain divergence and data selection bias~\cite{torralba2011unbiased,tommasi2017deeper}.

\ifextended
In fact, a recent work~\cite{laroca2022cross} showed that there are significant drops in \gls*{lpr} performance for most datasets when training and testing well-known \gls*{ocr} models (e.g., Facebook's Rosetta~\cite{borisyuk2018rosetta} and \trba~\cite{baek2019what}) in a \gls*{lodo} experimental setup. 
\else
In fact, a recent work~\cite{laroca2022cross} showed that there are significant drops in \gls*{lpr} performance for most datasets when training and testing well-known \gls*{ocr} models (e.g., \crnn and Facebook's Rosetta) in a \gls*{lodo} experimental setup. 
\fi
One might think that such disappointing results were caused by the fact that existing datasets for \gls*{lpr} are heavily biased towards specific regional identifiers~\cite{anagnostopoulos2006license,zhang2021robust_attentional}.
\ifalt
Nevertheless, the authors explored several data augmentation techniques to avoid overfitting, including one based on character permutation~\cite{goncalves2018realtime} that is known to successfully mitigate the impacts of such bias on the learning of recognition models~\cite{hochuli2018segmentation,goncalves2019multitask,laroca2021efficient}.
\else
Nevertheless, the authors explored several data augmentation techniques to avoid overfitting, including one based on character permutation~\cite{goncalves2018realtime} that is known to successfully mitigate the impacts of such bias on the learning of \gls*{lpr} models~\cite{goncalves2019multitask,laroca2021efficient,shashirangana2021license}.
\fi
Thus, we soon hypothesized that there are \emph{other} strong biases crept into \gls*{lpr} datasets.
This analysis is precisely what motivated our~work.
\section{Experiments}
\label{sec:experiments}

This section describes the experiments performed in this work.
We first list the datasets explored in our assessments, explaining why they were chosen and not others.
We also detail how the \gls*{lp} images from each dataset were selected and divided into training, validation and test subsets.
Then, we describe the \gls*{cnn} model employed for the dataset classification task (\emph{Name That Dataset!}) and provide implementation details.
Finally, we report the results achieved.

\subsection{Datasets}
\label{sec:experiments:datasets}

Our experiments were carried out on images from eight public \gls*{alpr} datasets introduced over the last decade:
\rodosolalpr~\cite{laroca2022cross}, \ssigsegplate~\cite{goncalves2016benchmark}, \ufop~\cite{mendesjunior2011towards}, \ufpralpr~\cite{laroca2018robust}, a reduced version of \ccpd~\cite{xu2018towards}, \chineselp~\cite{zhou2012principal}, \pku~\cite{yuan2017robust}, and \platesmaniaietcn~\cite{laroca2021efficient}.
The images of the first four datasets were acquired in three states of Brazil, while the images of the last four datasets were collected in various provinces of mainland~China. 
We cropped the \gls*{lp} regions from the original images (taken in urban environments) for our~experiments.

In this work, we chose to experiment with \glspl*{lp} from Brazil and mainland China because there are many \gls*{alpr} systems designed primarily for \glspl*{lp} from one of those regions~\cite{silva2017realtime,laroca2018robust,yang2018chinese,gong2022unified}.
Considering the objectives of our study, we also filter which \gls*{lp} images from each dataset to use in our experiments:
(i)~regarding the datasets collected in Brazil, we explore only \glspl*{lp} that have a single row of characters and gray as the background color; and 
(ii)~for the datasets acquired in mainland China, we explore only \glspl*{lp} that have a single row of characters and blue as the background color.
This protocol was adopted because the four datasets collected in each region have \glspl*{lp} with these characteristics.
In contrast, only some have \glspl*{lp} with other characteristics (e.g., the \ufop and \ssigsegplate datasets do not have any two-row \glspl*{lp}, and the \chineselp and \platesmaniaietcn datasets do not have any yellow \glspl*{lp}).
See \cref{tab:experiments:overview_datasets} for an overview of the datasets used in our experiments.
We labeled the color of each \gls*{lp} in every dataset to make this selection.
\iffinal
These annotations are publicly~available\customfootnote{\supplementaryAnnotations}.
\setcounter{fnAnnotations}{\thefncounter}
\else
These annotations will be made publicly~available.
\fi

\begin{table}[!htb]
\centering
\renewcommand{\arraystretch}{1.05}
\caption{The datasets used in our experiments.}
\label{tab:experiments:overview_datasets}

\vspace{-2mm}

\resizebox{0.9\linewidth}{!}{ 
\begin{tabular}{@{}lccl@{}}
\toprule
\textbf{Dataset} & \textbf{Year} & \textbf{\gls*{lp} Images} & \textbf{State / Province-City} \\ \midrule
\ufop~\cite{mendesjunior2011towards} & $2011$ & $244$ & Minas Gerais (BR) \\
\chineselp~\cite{zhou2012principal} & $2012$ & $400$ & Various (CN) \\
\ssigsegplate~\cite{goncalves2016benchmark} & $2016$ & $1{,}832$ & Minas Gerais (BR) \\
\pku~\cite{yuan2017robust} & $2017$ & $2{,}024$ & Anhui-Tongling (CN) \\
\ufpralpr~\cite{laroca2018robust} & $2018$ & $2{,}700$ & Paran\'{a} (BR) \\
\ccpd~\cite{xu2018towards} & $\phantom{\ast}2020^\ast$ & $\phantom{\dagger}25{,}000^\dagger$ & Anhui-Hefei (CN) \\
\platesmaniaietcn~\cite{laroca2021efficient} & $2021$ & $347$ & Various (CN) \\
\rodosolalpr~\cite{laroca2022cross} & $2022$ & $4{,}765$ & Esp\'{\i}rito Santo (BR) \\ \bottomrule \\[-2.2ex]
\multicolumn{4}{l}{\hspace{-2mm}$^\ast$\hspace{0.3mm}\scriptsize The \ccpd dataset was introduced in 2018 and last updated in 2020.} \\[-0.1ex]
\multicolumn{4}{l}{\hspace{-2mm}$^\dagger$\hspace{0.3mm}\scriptsize Following~\cite{liu2021fast}, we used a reduced version of \ccpd in our experiments.}
\end{tabular}
}
\end{table}

For reproducibility, it is essential to make clear how we divided the selected images from each of the datasets to train, validate and test the classification model (detailed in \cref{sec:experiments:model}).
The \ccpd, \rodosolalpr, \ssigsegplate and \ufpralpr datasets were split according to the protocols defined by the respective authors (i.e., the authors specified which images belong to which subsets), while the other datasets, which do not have well-defined evaluation protocols, were randomly split into $40$\% images for training; $20$\% images for validation; and $40$\% images for testing, following the split protocol adopted in the \ssigsegplate and \ufpralpr datasets\customfootnote{The training, validation, and test splits are available at \supplementarySplits}.
As the \ccpd dataset has many more images than the others (more than $350$K), we followed~\cite{liu2021fast} and performed our experiments using a reduced version with $25$K~images.

Three points should be noted.
First, for all datasets, we were careful not to have images of the same \gls*{lp} in different subsets (otherwise, different images of an \gls*{lp} could appear in both the training and test sets, for example).
Second, as the chosen datasets have different numbers of test images, we randomly sample a set of $N$ test set images from different datasets to predict which dataset each image belongs to (for each region, $N$ is constrained by the smallest number of images in the test sets).
This experiment is repeated $100$ times with diﬀerent splits and we report the average results.
Similar protocols were adopted in~\cite{torralba2011unbiased,khosla2012undoing,tommasi2017deeper}.
\ifalt
Third, we used Albumentations~\cite{albumentations}, a well-known library for image augmentation, to balance the number of training images from different datasets, thus avoiding overfitting~\cite{krinski2022light}.
\else
Third, we used Albumentations~\cite{albumentations}, a well-known library for image augmentation, to balance the number of training images from different datasets, thus avoiding overfitting.
\fi
Transformations applied to generate new images include, but are not limited to, random noise, random JPEG compression, random shadows, and random perturbations of hue, saturation and brightness.

\begin{figure}[!htb]
    \centering
    \captionsetup[subfigure]{captionskip=1.25pt} 
    
    \resizebox{0.85\linewidth}{!}{
    \includegraphics[width=0.26\linewidth]{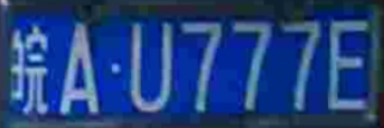} \hspace{-1.75mm}
    \includegraphics[width=0.26\linewidth]{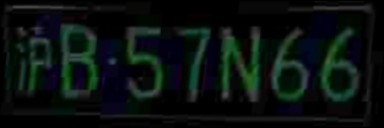} \hspace{-1.75mm}
    \includegraphics[width=0.26\linewidth]{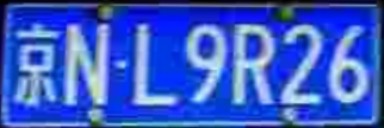}
    }
    
    \vspace{0.7mm}

    \resizebox{0.85\linewidth}{!}{
    \includegraphics[width=0.26\linewidth]{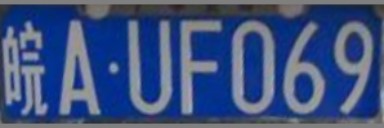} \hspace{-1.75mm}
    \includegraphics[width=0.26\linewidth]{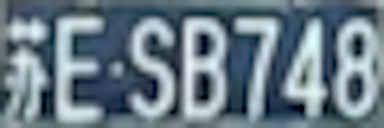} \hspace{-1.75mm}
    \includegraphics[width=0.26\linewidth]{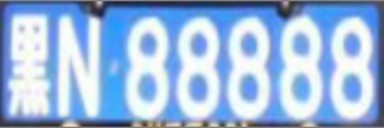}
    }
    
    \vspace{0.7mm}
    
    \resizebox{0.85\linewidth}{!}{
    \includegraphics[width=0.26\linewidth]{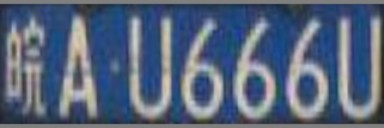} \hspace{-1.75mm}
    \includegraphics[width=0.26\linewidth]{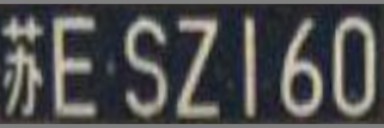} \hspace{-1.75mm}
    \includegraphics[width=0.26\linewidth]{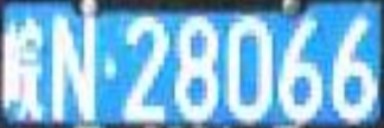}
    }
    
    \vspace{0.7mm}
    
    \resizebox{0.85\linewidth}{!}{
    \includegraphics[width=0.26\linewidth]{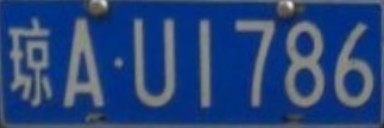} \hspace{-1.75mm}
    \includegraphics[width=0.26\linewidth]{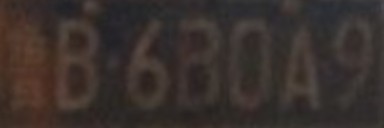} \hspace{-1.75mm}
    \includegraphics[width=0.26\linewidth]{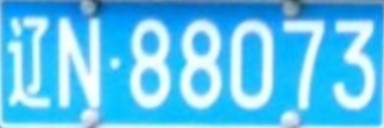}
    }
    
    \vspace{-2mm}
    
    \caption{Some Chinese \glspl*{lp} from the datasets used in our experiments. From top to bottom: \ccpd~\cite{xu2018towards}, \chineselp~\cite{zhou2012principal}, \pku~\cite{yuan2017robust} and \platesmaniaietcn~\cite{laroca2021efficient}.
    The first character on each \gls*{lp} is a Chinese character representing the province in which the vehicle is affiliated.
    The second character is an English letter representing the city --~in the province~-- in which the vehicle is affiliated~\cite{laroca2021efficient}.
    }
    \label{fig:samples-chinese-lps}
\end{figure}

Hereinafter, ``Brazilian \glspl*{lp}'' refer to gray single-row \glspl*{lp} from vehicles registered in Brazil (prior to the adoption of the Mercosur layout), and ``Chinese \glspl*{lp}'' refer to blue single-row \glspl*{lp} from vehicles registered in mainland~China.
While some examples of Brazilian \glspl*{lp} can be seen in \cref{fig:name-that-dataset}, some Chinese \glspl*{lp} from the chosen datasets are shown in \cref{fig:samples-chinese-lps}.

One may have noticed that all \gls*{lp} images we showed (both in Fig. 1 and Fig. 2) are quite horizontal, tightly bounded, and ``easy'' to read.
This is because we rectified all \gls*{lp} images to eliminate biases such as repetitive tilt angles caused by specific camera positions in images from distinct datasets.
This procedure (i.e., \gls*{lp} rectification) has been performed quite frequently in the literature, as it makes the recognition methods considerably more robust to distortions caused by oblique views~\cite{weihong2020research,silva2022flexible,wang2022rethinking}.
To perform the rectification, we labeled the position ($x$, $y$) of the four corners of each \gls*{lp} in the eight datasets that do not contain such labels (only the \ccpd and \rodosolalpr datasets have corner annotations for all \glspl*{lp}).
\iffinal
These annotations are also publicly available\footnotemark[\thefnAnnotations].
\else
These annotations will also be made publicly~available.
\fi

\subsection{Classification Model}
\label{sec:experiments:model}

For the dataset classification task (\emph{Name That Dataset!}), we designed a lightweight \gls*{cnn} architecture called \model.
\iffinal
It is inspired by the \cdcc model~\cite{laroca2021towards} and is relatively similar to the model used for this same task in~\cite{mclaughlin2015data}.
\else
It is inspired by the network proposed in [hidden for review] and is relatively similar to the model used for this same task in~\cite{mclaughlin2015data}.
\fi

\model's architecture is shown in Table~\ref{tab:architecture}.
As can be seen, the model is relatively shallow, with three convolutional layers containing $16$/$32$/$64$ filters, each followed by a max-pooling layer with a $2\times2$~kernel and stride~$=2$.
Batch normalization, followed by a \gls*{relu}, is added after each convolutional layer.
We evaluated several changes to this architecture, such as using depthwise separable convolutional layers, convolutional layers with stride~=~$2$ (removing the max-pooling layers), and different input sizes and numbers of filters.
However, better results were not obtained (we conducted these experiments in the validation~set).

\begin{table}[!htb]
\centering
\caption{\model's layers and hyperparameters.}
\label{tab:architecture}

\vspace{-2mm}

\resizebox{0.85\linewidth}{!}{ %
\begin{tabular}{@{}cccccc@{}}
\toprule
\textbf{\#} & \textbf{Layer} & \textbf{Filters} & \textbf{Size / Stride} & \textbf{Input} & \textbf{Output} \\ \midrule
$0$ & conv & $16$ & $3 \times 3 / 1$ & $192 \times 64 \times 3$ & $192 \times 64 \times 16$ \\
$1$ & max &  & $2 \times 2 / 2$ & $192 \times 64 \times 16$ & $96 \times 32 \times 16$ \\
$2$ & conv & $32$ & $3 \times 3 / 1$ & $96 \times 32 \times 16$ & $96 \times 32 \times 32$ \\
$3$ & max &  & $2 \times 2 / 2$ & $96 \times 32 \times 32$ & $48 \times 16 \times 32$ \\
$4$ & conv & $64$ & $3 \times 3 / 1$ & $48 \times 16 \times 32$ & $48 \times 16 \times 64$ \\
$5$ & max & & $2 \times 2 / 2$ & $48 \times 16 \times 64$ & $24 \times 8 \times 64$ \\
$6$ & flatten & & & $24 \times 8 \times 64$ & $12288$ \\

\midrule
\textbf{\#} & \textbf{Layer} & & \textbf{Units} & \textbf{Input} & \textbf{Output} \\ \midrule
$7$ & dense & & $128$ & $12288$ & $128$ \\
$8$ & dense & & $4$ & $128$ & $4$ \\ \bottomrule
\end{tabular}
} %
\end{table}

\ifextended
The \model model was implemented using Keras\customfootnote{\url{https://keras.io/}}.
\else
The \model model was implemented using Keras.
\fi
We used the Adam optimizer, initial learning rate~=~$10$\textsuperscript{-$3$} (with \textit{ReduceLROnPlateau}'s patience = $3$ and factor~=~$10$\textsuperscript{-$1$}), batch size~=~$64$, max epochs~=~$50$, and patience~=~$7$ (the number of epochs with no improvement after which training is stopped).

All experiments were performed on a computer with an AMD Ryzen Threadripper $1920$X $3.5$GHz CPU, $96$~GB of RAM, and an NVIDIA Quadro RTX~$8000$ GPU~($48$~GB). 
In this setup, \model runs at $\approx720$ FPS (using batch size~=~$1$). 

\subsection{Results}
\label{sec:experiments:results}

\begin{figure}[!htb]
    \centering
    \resizebox{\linewidth}{!}{                      \includegraphics[width=\linewidth]{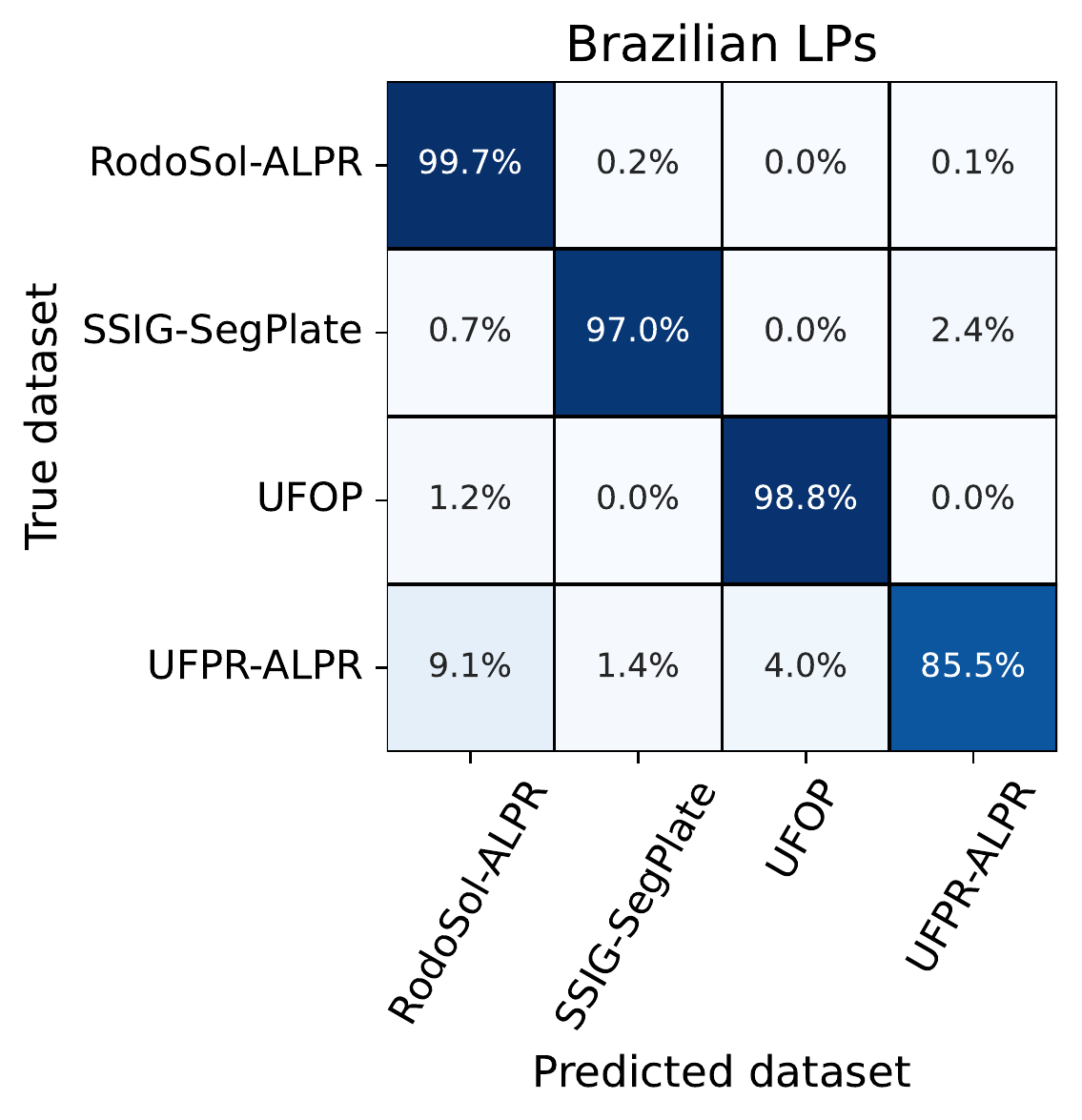} \quad
        \includegraphics[width=\linewidth]{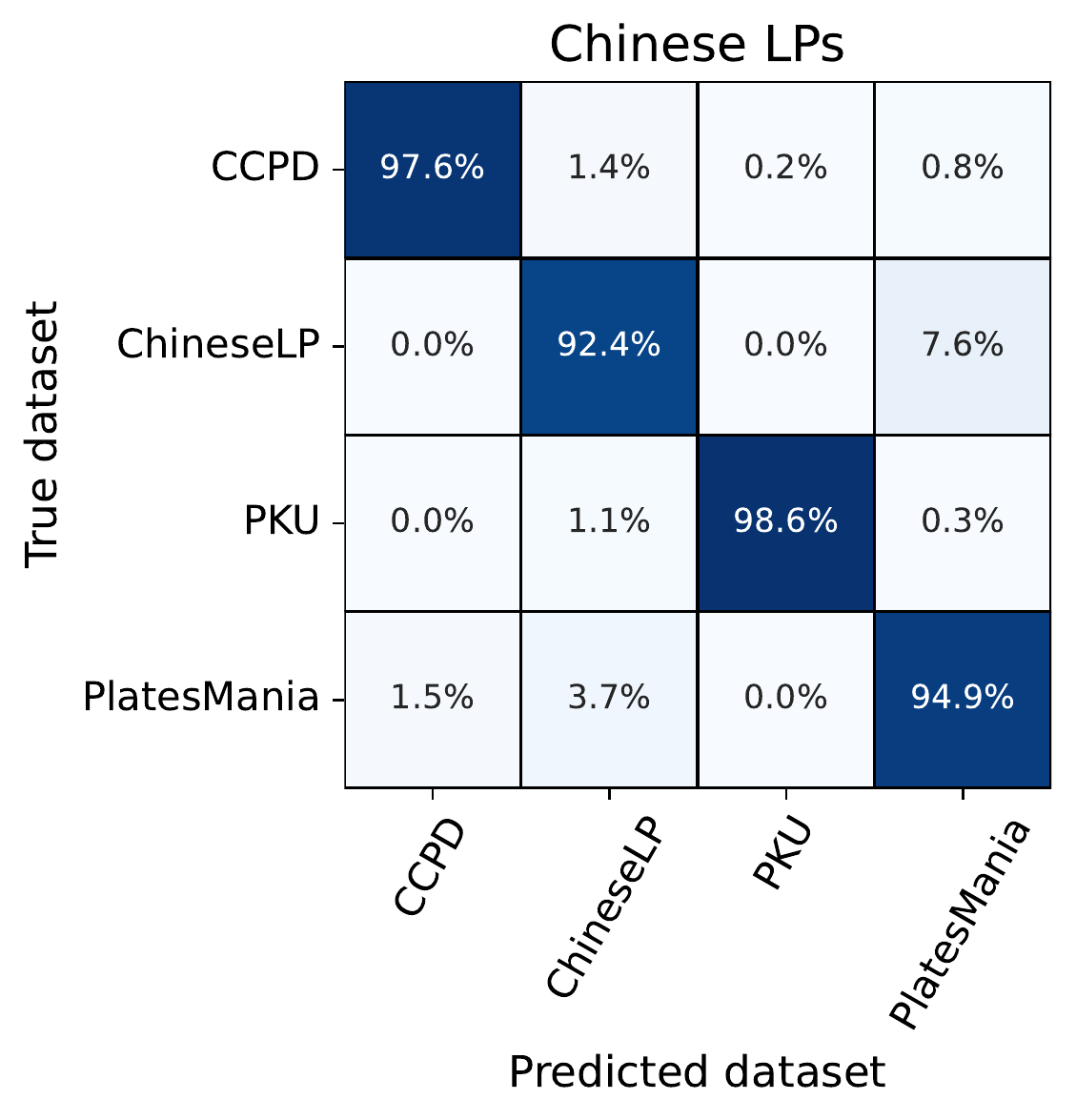}
    }
    
    \vspace{-3.5mm}
    
    \caption{Confusion matrices for a classifier (\model) trained to predict the source dataset of a given \gls*{lp} image; left: Brazilian \glspl*{lp}, right: Chinese \glspl*{lp}.
    }
    \label{fig:confusion_matrices}
\end{figure}

In this subsection, we report the results obtained by \model in the dataset classification task (\emph{Name That Dataset!}).
\cref{fig:confusion_matrices} shows the confusion matrices for Brazilian~(left) and Chinese~(right) \glspl*{lp}.
There is a clearly pronounced diagonal in both matrices, indicating that \emph{each dataset does have a unique, identifiable ``signature''}; it is worth noting that only about $25$\% accuracy would be expected if the classifier was operating at chance levels, as would happen if the \gls*{lp} images from each dataset were fully unbiased samples.
The overall accuracy was $\bracc$\% for Brazilian \glspl*{lp} and $\cnacc$\% for Chinese~\glspl*{lp}.

\begin{figure}[!htb]
    \centering
    \captionsetup[subfigure]{captionskip=1.5pt} 
    
    \subfloat[\rodosolalpr (MSE = $174$)]{
    \includegraphics[width=0.45\linewidth]{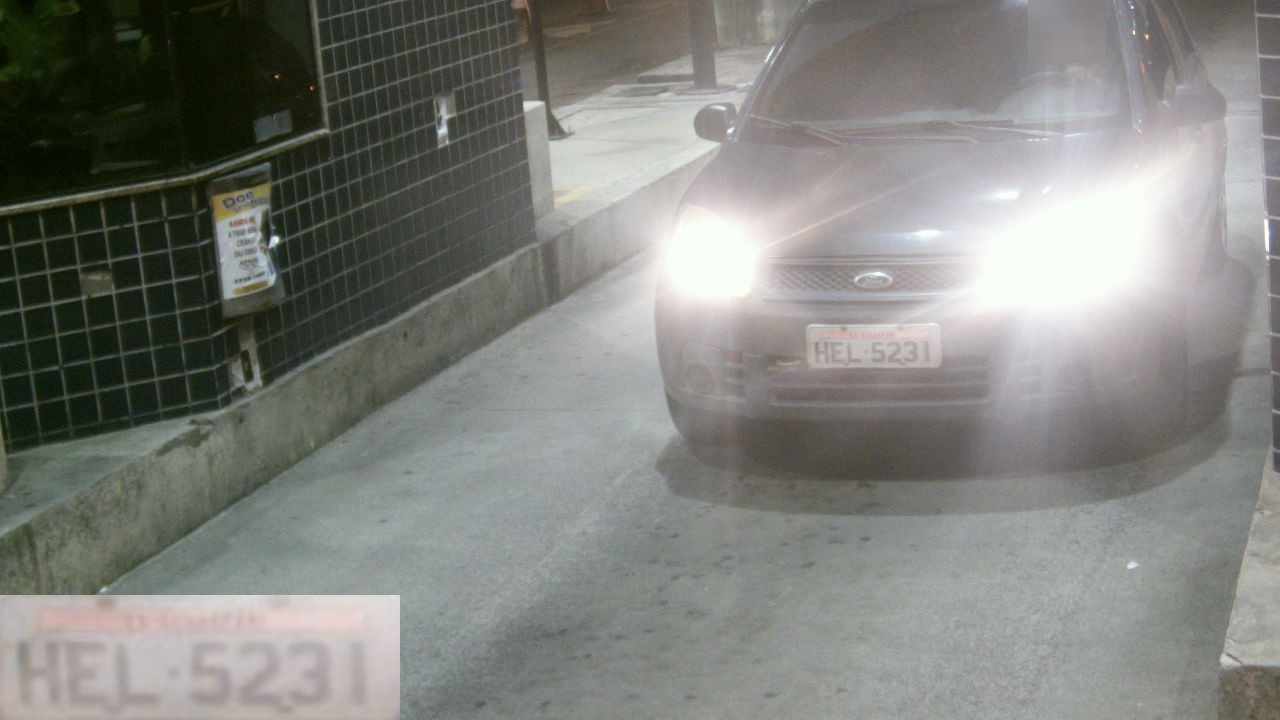}
    \includegraphics[width=0.45\linewidth]{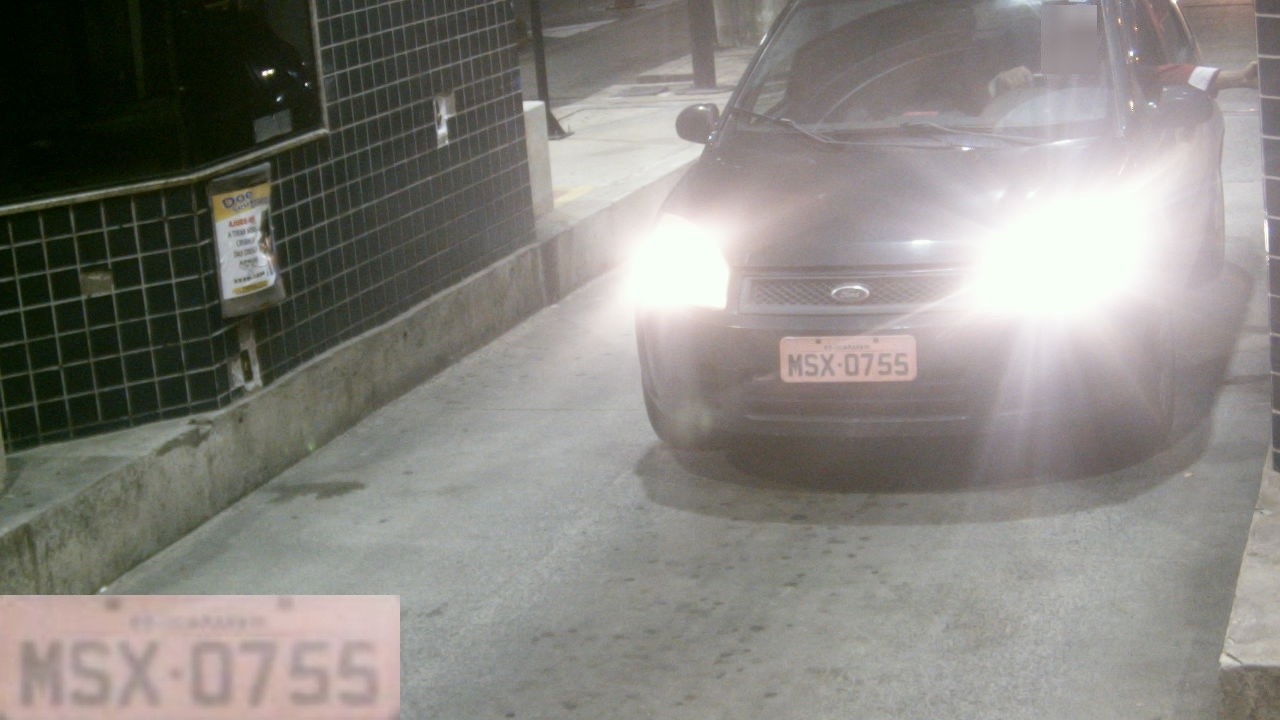}
    }
    
    \vspace{1.75mm}
    
    \subfloat[\rodosolalpr (MSE = $407$)]{
    \includegraphics[width=0.45\linewidth]{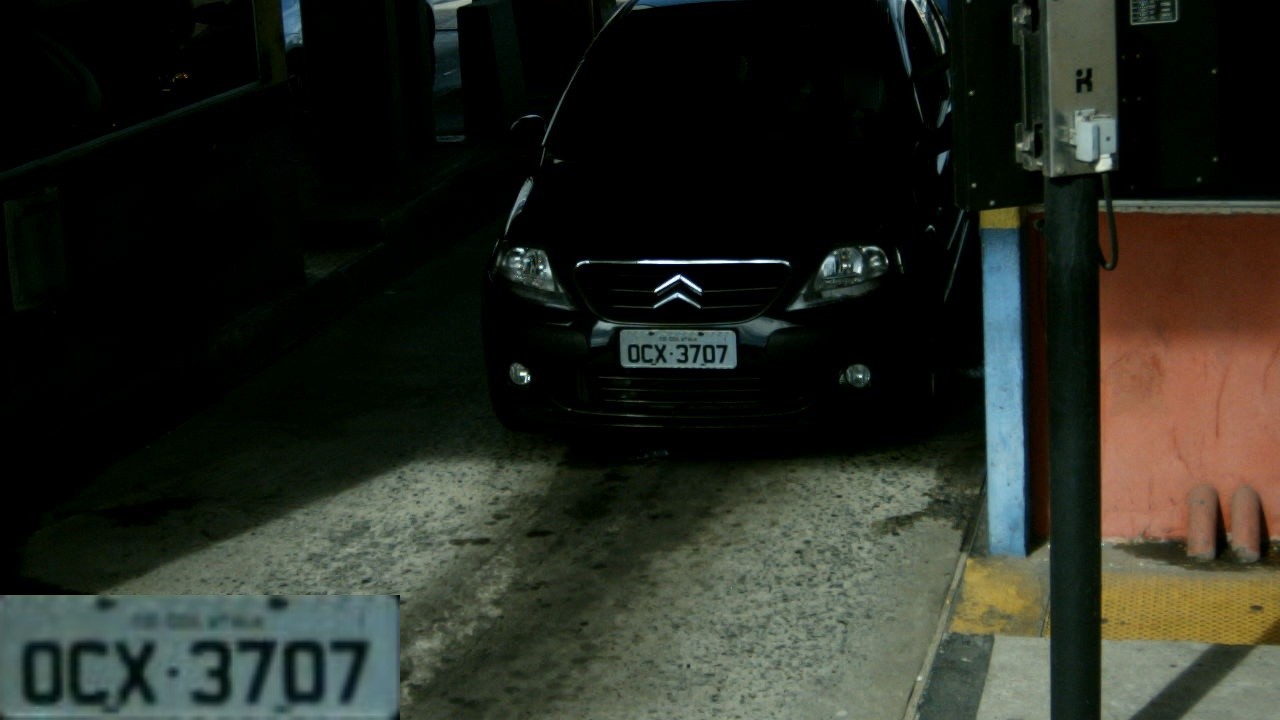}
    \includegraphics[width=0.45\linewidth]{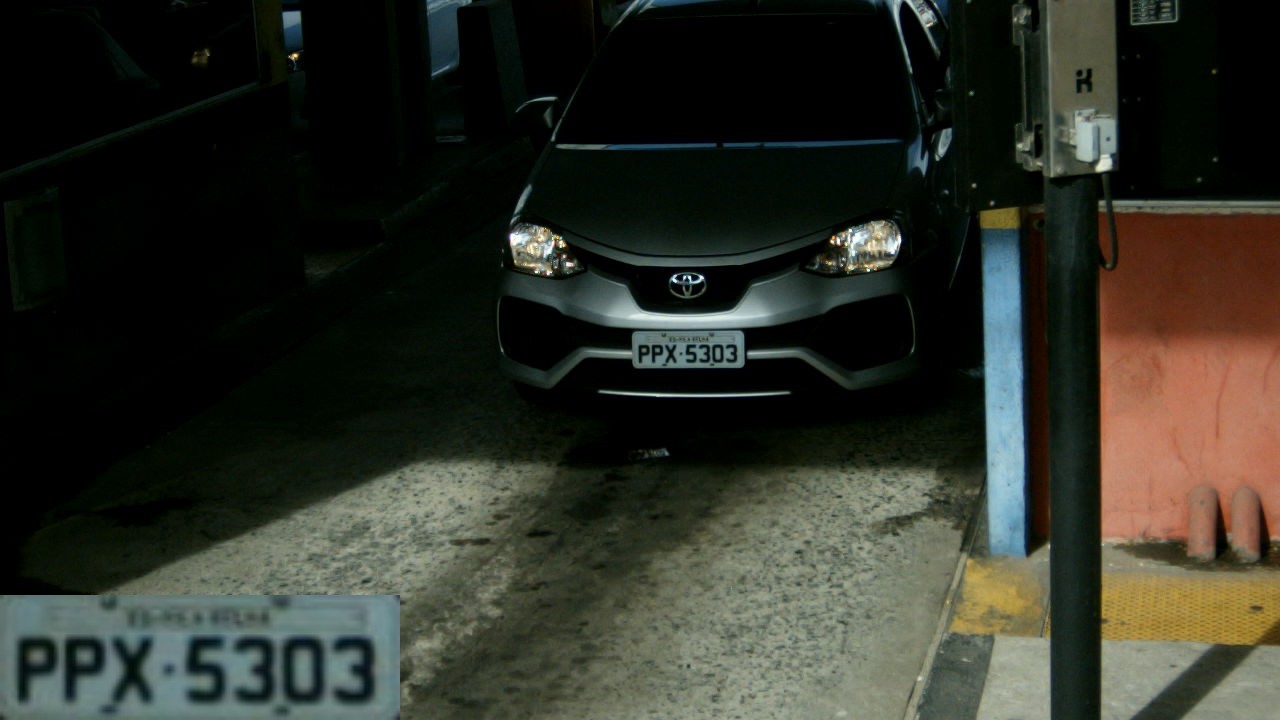}
    }
    
    \vspace{1.75mm}
    
    \subfloat[\ufpralpr (MSE = $1{,}686$)]{
    \includegraphics[width=0.45\linewidth]{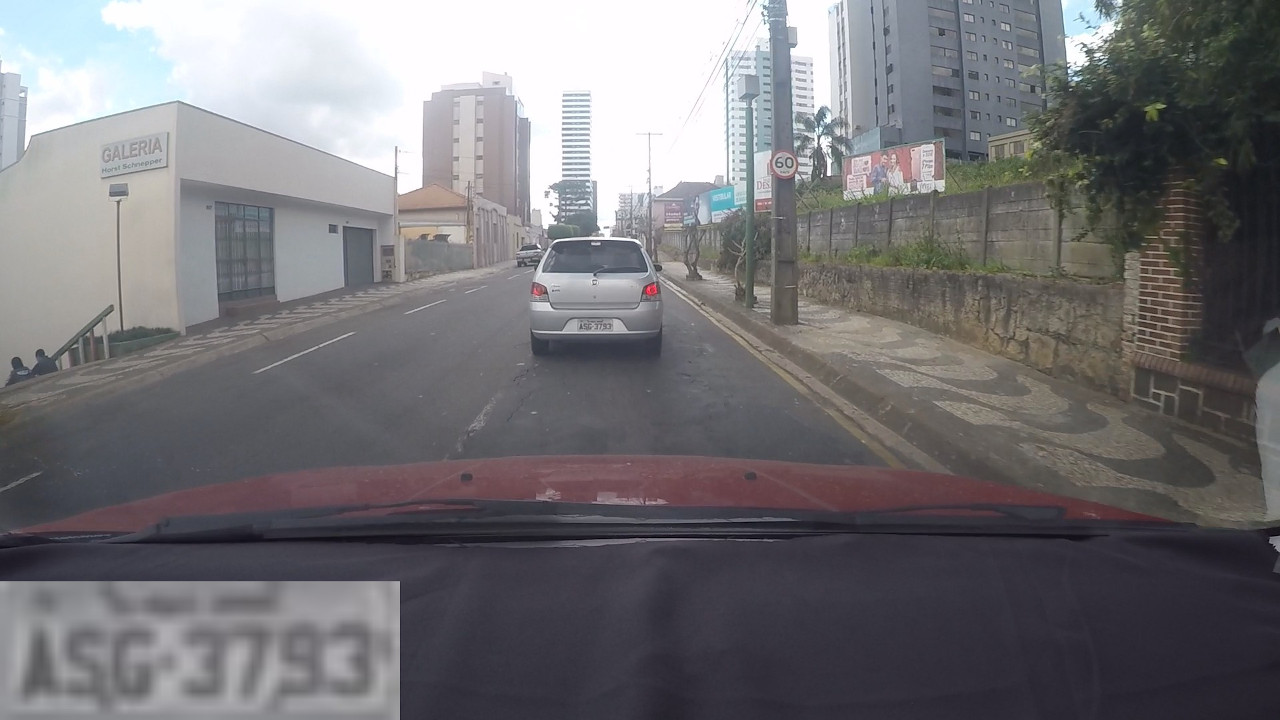}
    \includegraphics[width=0.45\linewidth]{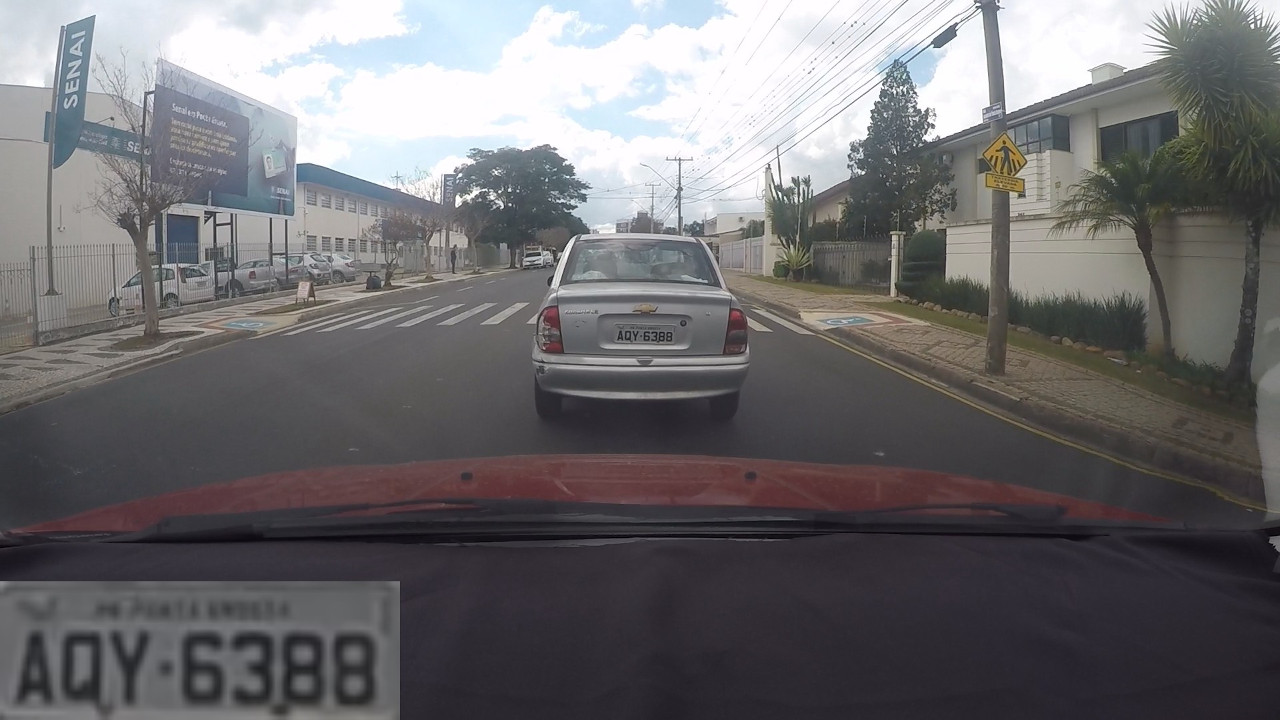}
    }
    
    \vspace{1.75mm}
    
    \subfloat[\ufpralpr (MSE = $1{,}700$)]{
    \includegraphics[width=0.45\linewidth]{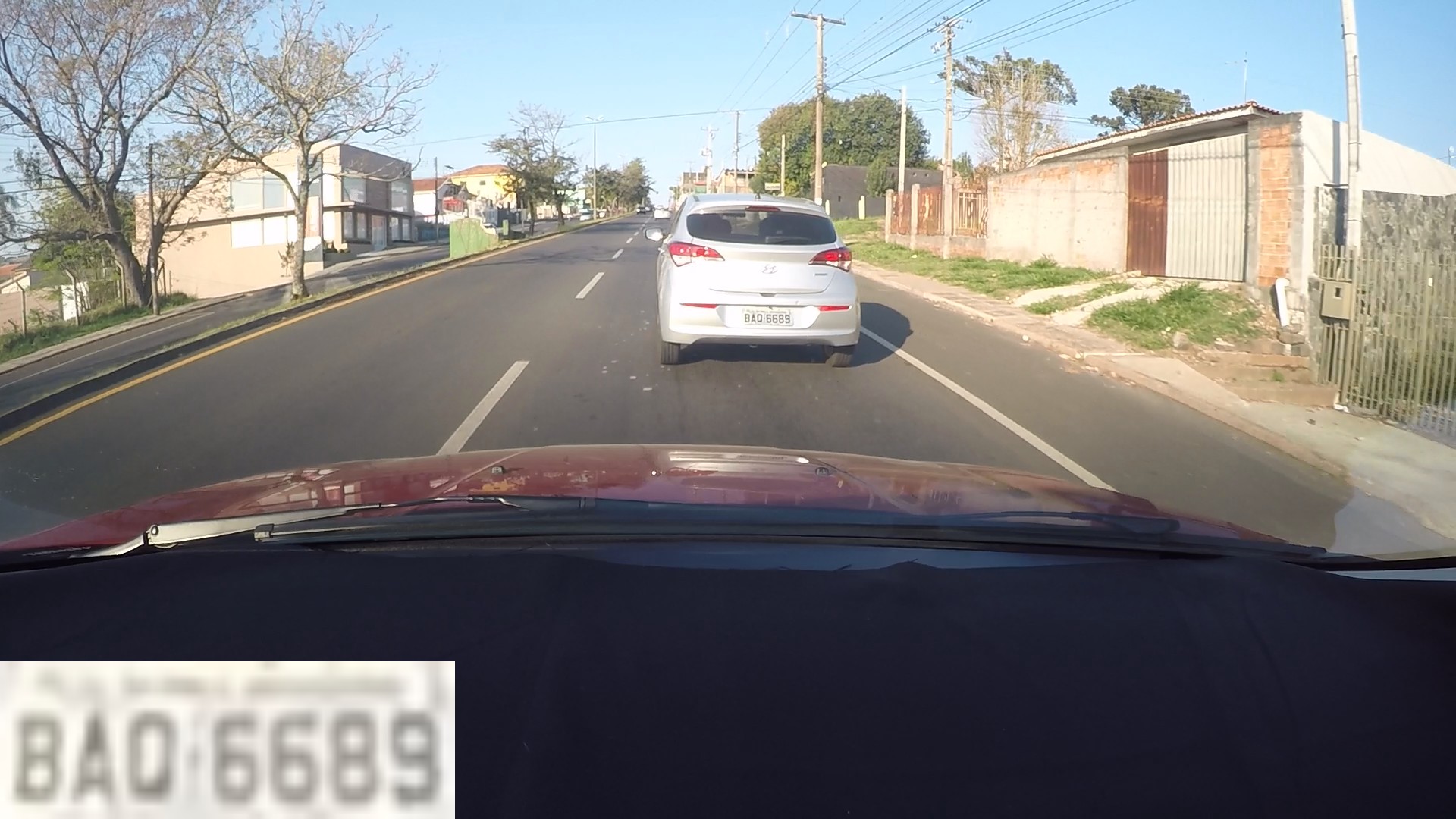}
    \includegraphics[width=0.45\linewidth]{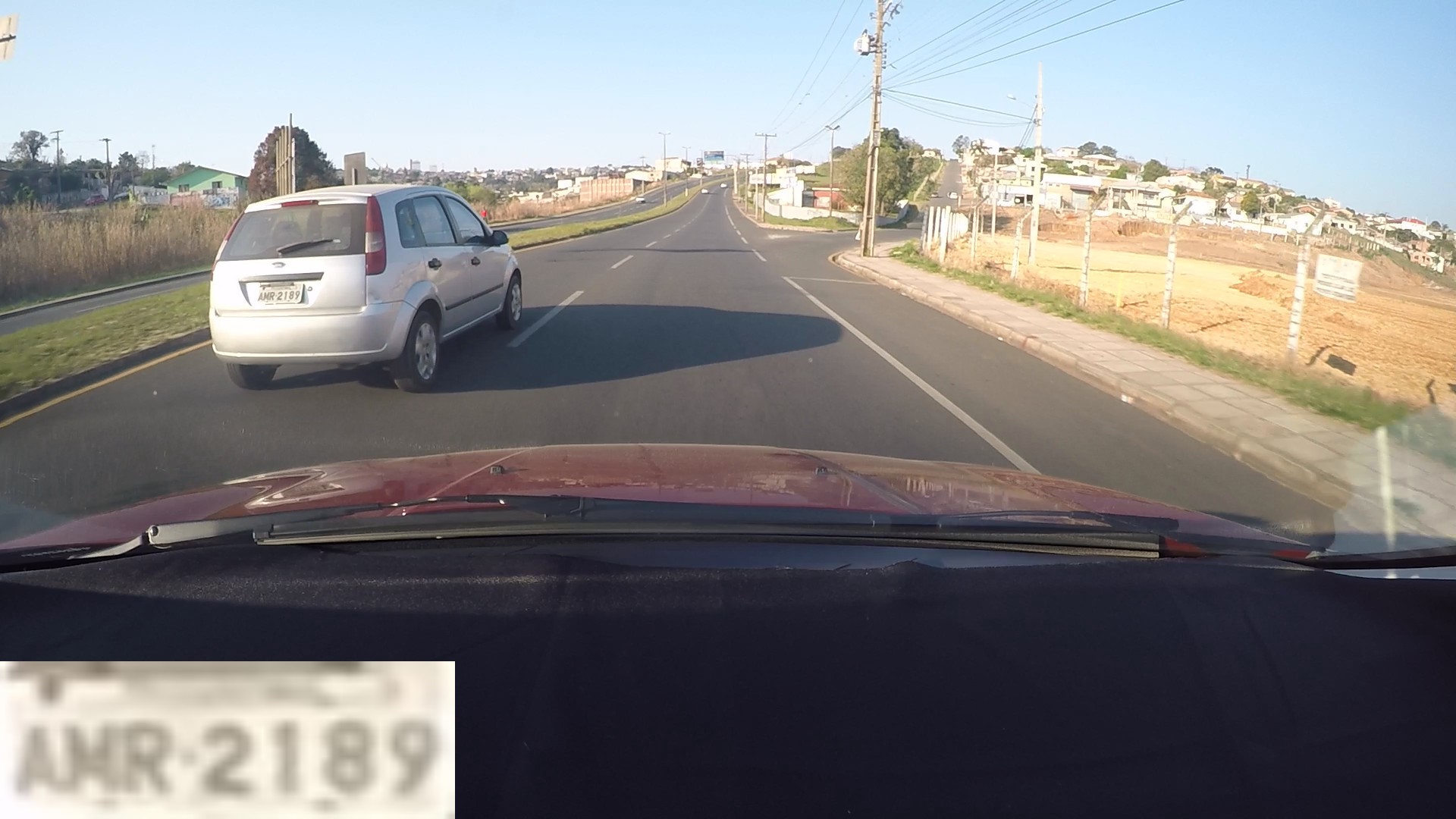}
    }
    
    \caption{Two pairs of the most similar images (in terms of \gls*{mse}) from distinct subsets from each of the \rodosolalpr~(a, b) and \ufpralpr~(c, d) datasets.
    In each pair, the left image belongs to the training set, while the right one belongs to the test set. 
    Observe that \glspl*{lp} from different images captured by static cameras may be quite resembling.
    We show a zoomed-in version of the \gls*{lp} in the lower left region of each image for better viewing.}
    \label{fig:similar-images-training-testing}
\end{figure}

The results show that the \model model is more successful in classifying \gls*{lp} images from the datasets acquired with static cameras  (\rodosolalpr, \ssigsegplate, \ufop, and \pku) than \gls*{lp} images from the datasets captured by handheld (\ccpd, \chineselp, and \platesmaniaietcn) or moving cameras (\ufpralpr).
We believe this is because images collected by static cameras have many characteristics in common, not just the background.
These similarities are probably present to some extent in the \gls*{lp} regions, explaining the greater accuracy reached by the network in such images.
To illustrate, in \cref{fig:similar-images-training-testing}, we show two pairs of the most similar images --~in terms of \gls*{mse}~-- from distinct subsets from each of the \rodosolalpr and \ufpralpr datasets (the datasets where the highest and worst accuracy were achieved, respectively).
Observe that factors common in images taken by static cameras, such as similar position and distance of the vehicles to the camera, may cause the \glspl*{lp} from different images to be quite resembling (note that this is not always the case; it may seem so because we focused on the \emph{most} similar pairs of images from these datasets for this~analysis).

It is intuitive to suppose that the network may simply have learned the most frequent regional characters in each dataset (e.g., most \glspl*{lp} in the \ccpd dataset have~\begin{CJK*}{UTF8}{gbsn}`皖\end{CJK*}' as the first character).
However, this does not hold since \model correctly classified more than $97$\% of the \gls*{lp} images from both datasets collected in the Brazilian state of Minas Gerais (\ssigsegplate and \ufop) and from both datasets acquired in the Anhui province in mainland China (\ccpd and \pku).

By carefully analyzing the confusion matrices in \cref{fig:confusion_matrices}, we noticed that almost all incorrect predictions on Chinese \glspl*{lp} were between the \chineselp and \platesmaniaietcn datasets.
We consider this occurred because both datasets have images collected from the internet (the other six datasets do not contain any images from the internet).
Specifically, all images from the \platesmaniaietcn datasets were downloaded from the internet~\cite{laroca2021efficient}, while $\approx39$\% of the \chineselp's images were taken from the internet~\cite{zhou2012principal}.
It makes perfect sense that the bias is less pronounced when the images come from multiple sources.
The classifier still managing to achieve high accuracy rates in both datasets is due to \emph{selection bias}~\cite{torralba2011unbiased,mclaughlin2015data,wachinger2021detect}, which arises because authors building a dataset select images with specific purposes in mind, thus reducing the variability of the data (in many cases without even realizing it).
Furthermore, these datasets have images with different quality levels, as they were introduced years apart and the capture devices evolved considerably in the time between them being~collected.

\begin{figure}[!htb]
    \centering
    
    \resizebox{\linewidth}{!}{
    \includegraphics[width=0.48\linewidth]{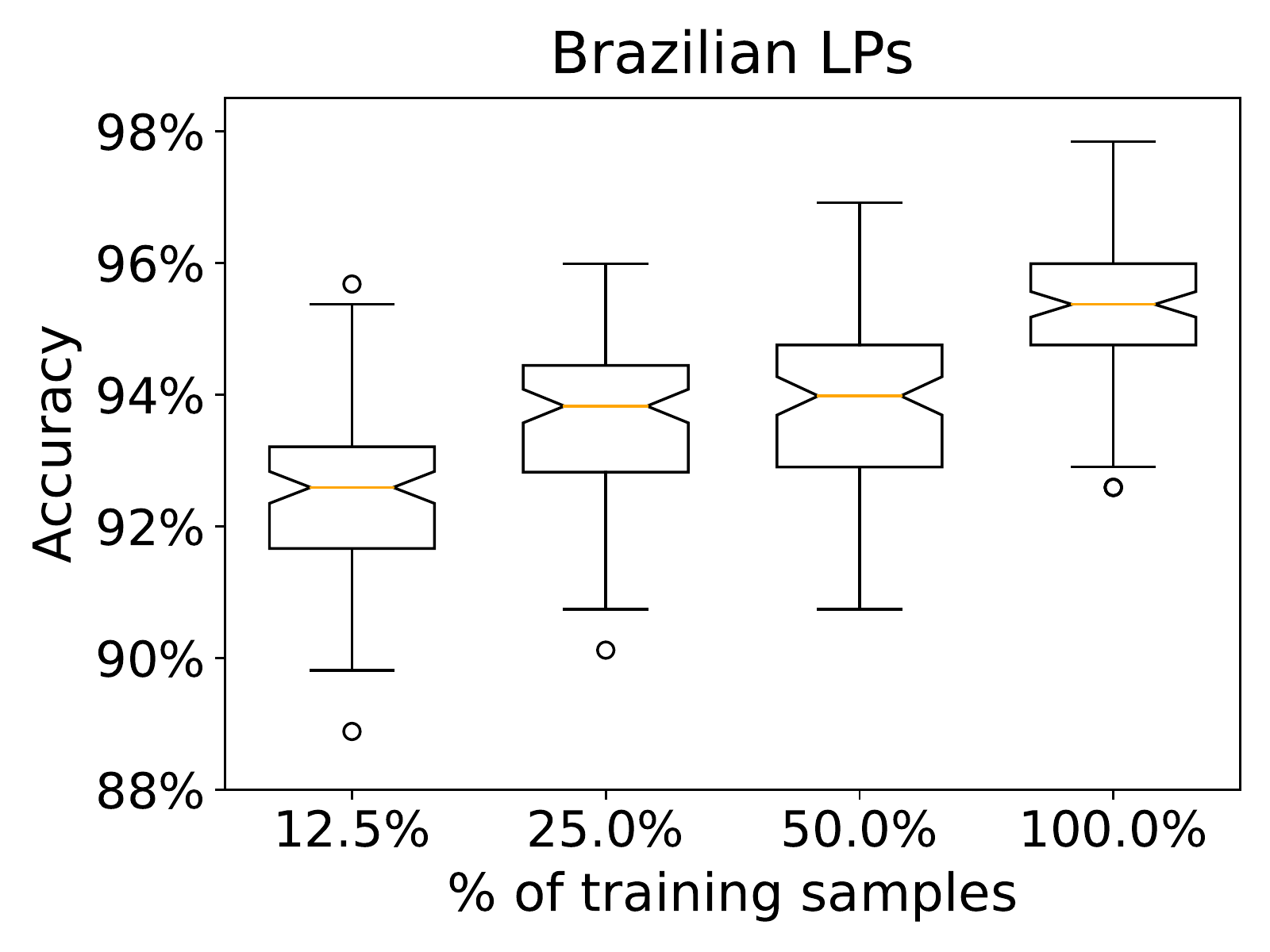} \,
    \includegraphics[width=0.48\linewidth]{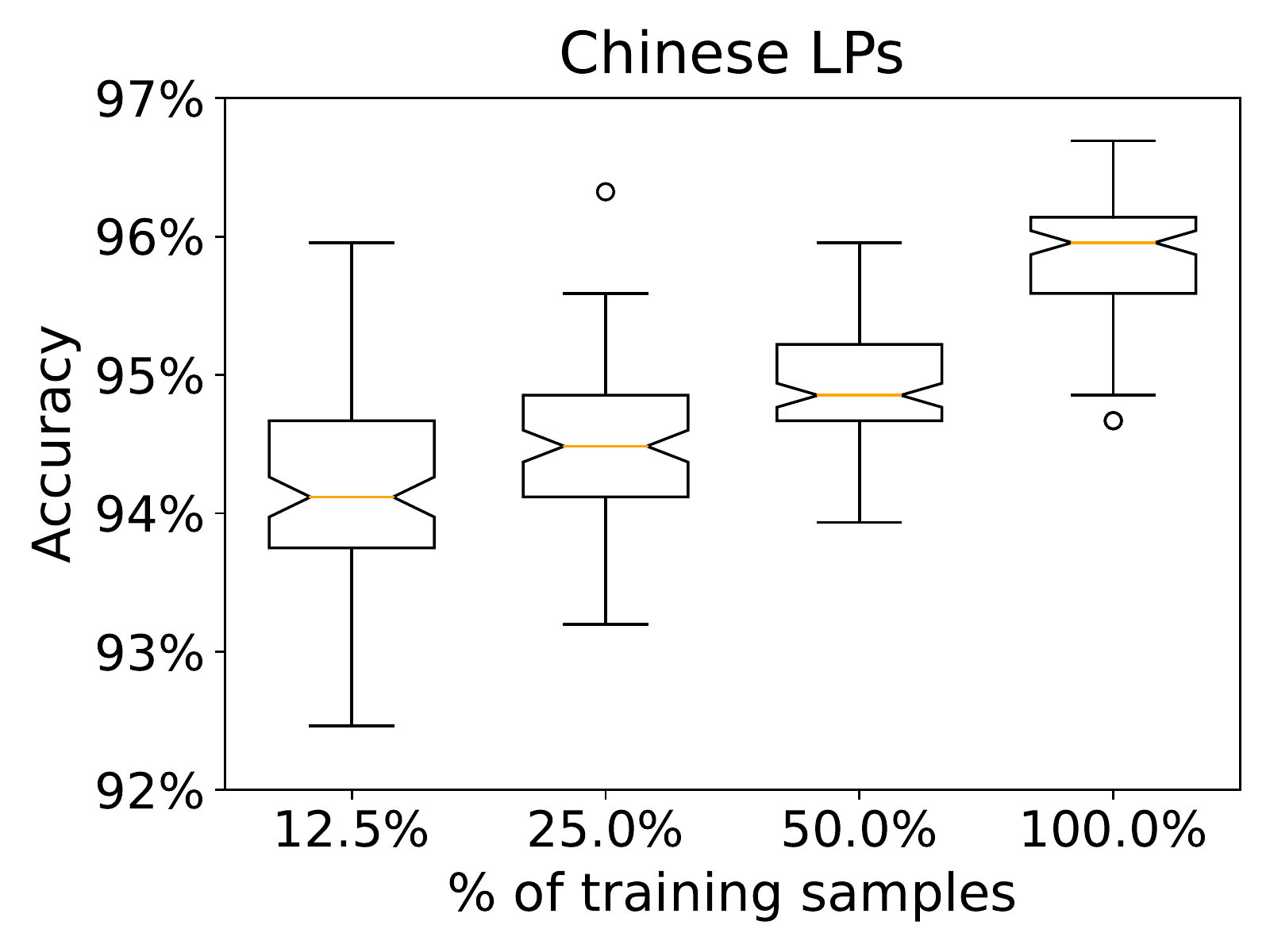}
    }
    
    \vspace{-3.5mm}
    
    \caption{Classification performance as a function of training data size. The performance does not seem to be saturated for either Brazilian or Chinese~\glspl*{lp}.}
    \label{fig:boxplots}
\end{figure}

An interesting point Torralba et al.~\cite{torralba2011unbiased} highlighted was that by using more training data, the classification accuracy could be increased with no immediate signs of saturation.
Intrigued by such results, we trained the \model model three more times for each \gls*{lp} layout: using $50$\%, $25$\% and $12.5$\% of the training data (randomly selected).
As shown in \cref{fig:boxplots}, the same behavior is observed in our experiments, i.e., the accuracy consistently improves as the size of the training set~increases.

\begin{figure}[!htb]
    \centering

    \resizebox{\linewidth}{!}{
    \includegraphics[width=0.48\linewidth]{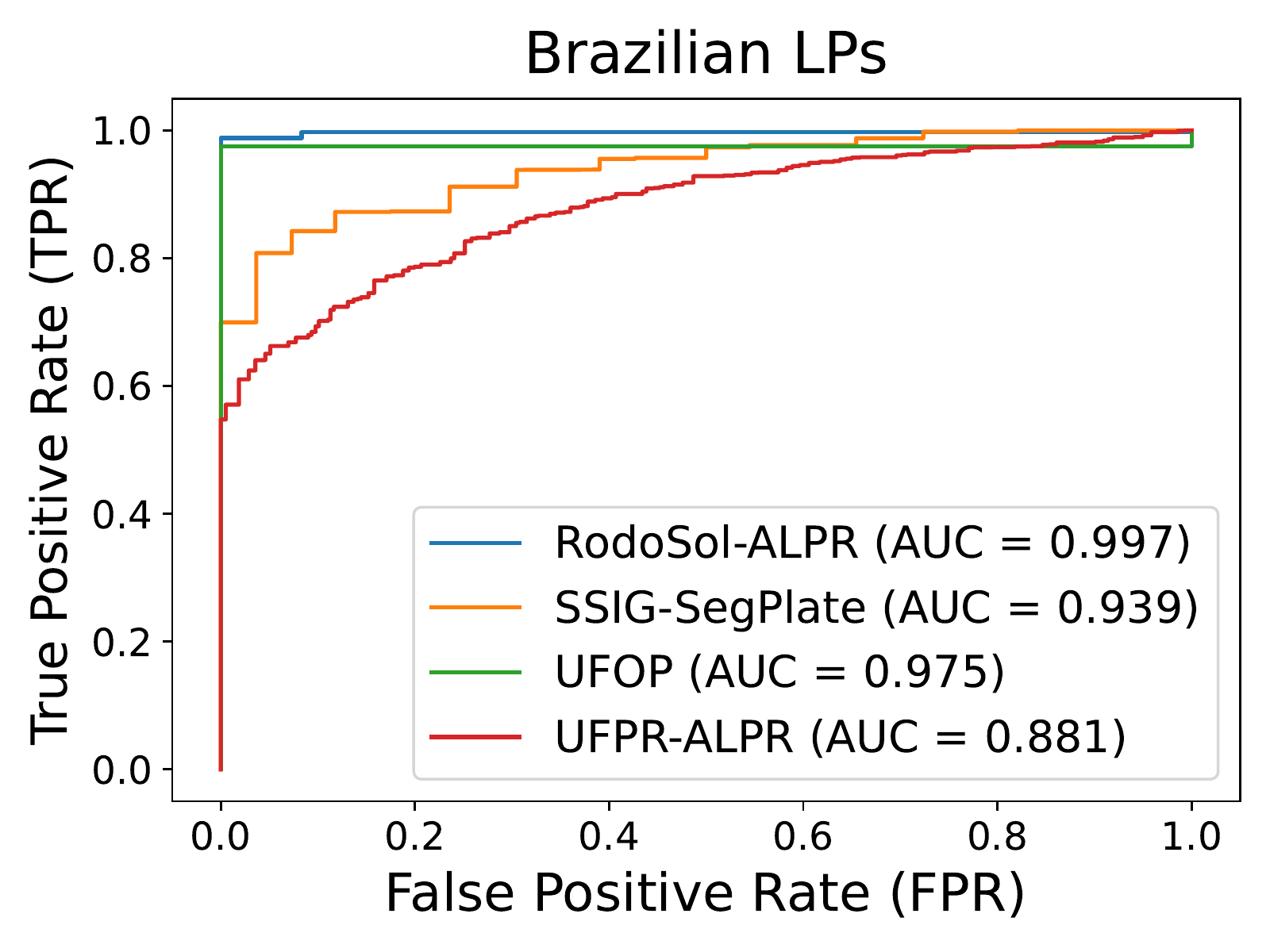} \,
    \includegraphics[width=0.48\linewidth]{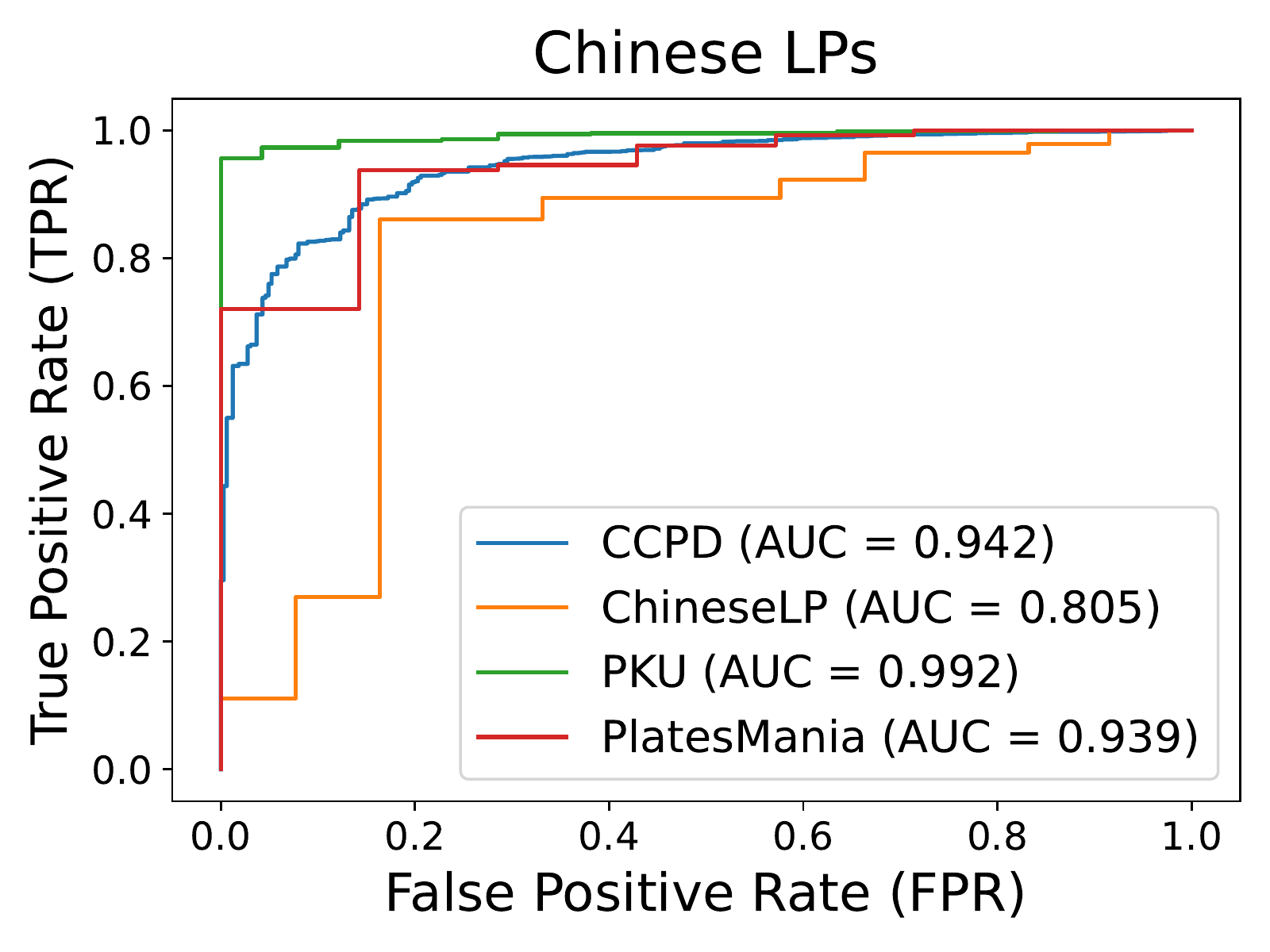}
    }
    
    \vspace{-3.5mm}
    
    \caption[ignore]{\gls*{roc} curves for Brazilian and Chinese \glspl*{lp}. Note the high \acrfull*{auc} values, which indicate that \model performs considerably well at distinguishing between \gls*{lp} images from different datasets.}
    \label{fig:roc-curves}
\end{figure}

Another interesting finding is that the classifier predicts the source dataset correctly with a significantly higher confidence value than when it predicts incorrectly.
The mean confidence values for correctly classified Brazilian and Chinese \glspl*{lp} were $98.5$\% and $98.1$\%, respectively, while the mean confidence values for incorrectly classified Brazilian and Chinese \glspl*{lp} were $79.7$\% and $74.3$\%, respectively.
\cref{fig:roc-curves} shows the \gls*{roc} curves for Brazilian~(left) and Chinese~(right) \glspl*{lp}.
\ifextended
Since \gls*{roc} curves are typically used in binary classification, we binarized the classifier's output (per class) to draw one \gls*{roc} curve per~dataset.
\else
\fi

\section{Discussion}
\label{sec:discussion}

Considering that the \model model --~which is relatively shallow~-- can predict the source dataset of an \gls*{lp} image with accuracy above $\bothAccMore$\%, we conjecture that most \gls*{lpr} models --~which are considerably deeper~-- are actually learning and exploiting such signatures to improve the results achieved in seen datasets at the cost of losing generalization capability.
The intuition behind this conjecture is as follows: consider the \ssigsegplate dataset~\cite{goncalves2016benchmark} as an example, it has many \glspl*{lp} with the letter~`O' as the first character but no \gls*{lp} with the letter~`Q' in that position.
Hence, an \gls*{lpr} model capable of identifying that a given \gls*{lp} image belongs to this dataset may predict the letter `O' as the first character even if the character looks more like `Q' than `O' due to noise or other factors.
However, the relatively high recognition rates achieved in the \ssigsegplate dataset would likely not be reached in unseen~datasets.

While we are unaware of previous attempts to \emph{Name That Dataset!} in the \gls*{lpr} context, our findings echo the concerns recently raised by Laroca et al.~\cite{laroca2022cross}.
They evaluated the cross-dataset generalization of $12$ well-known \gls*{ocr} models applied to \gls*{lpr} on nine datasets.
As mentioned in \cref{sec:related_work}, they observed significant drops in performance for most datasets when training and testing the models in a \gls*{lodo} fashion (i.e., one dataset is selected as the test set, and the remaining datasets are used for training).
Interestingly, the authors drew attention to the fact that the errors under the \gls*{lodo} protocol did not occur in challenging cases and thus were probably caused by \emph{differences in the training and test~images}.

We believe that the main cause of dataset bias is related to the cameras used to collect the images in each dataset.
Taking the results achieved in Brazilian \glspl*{lp} as an example, the lowest accuracy (i.e., less pronounced bias) was reported for the \ufpralpr dataset, which was captured by three non-static cameras of different price ranges.
In contrast, the other datasets have images acquired by a single static camera (\ssigsegplate and \ufop) or by multiple static cameras of the same model (\rodosolalpr).
In the same direction, another probable cause of bias relates to how the images were stored in different datasets.
For example, the \ccpd dataset contains highly compressed images~\cite{laroca2022cross,silva2022flexible} while most other datasets do not.
\model probably exploited the detection of artifacts in the highly compressed \gls*{lp} images for better~classification.

Some works have related dataset bias to image backgrounds~\cite{mclaughlin2015data,tian2018eliminating}.
For example, a classifier may be classifying images of the class ``boat'' without focusing on the boat itself but rather on the water below or the shore in the distance~\cite{torralba2011unbiased}.
Although we are convinced that we have eliminated such bias by performing our experiments on rectified \gls*{lp}~images, it is worth noting that the corner annotations provided along with the \ccpd dataset~\cite{xu2018towards} are not as accurate as those we made or those from the \rodosolalpr dataset~\cite{laroca2022cross}.
The \model model may have exploited these subtle distinctions as~well.

It is worth noting that while these conclusions have been reached for the particular classifier used in our experiments, similar trends are expected to hold for similar models~\cite{mclaughlin2015data}.

We consider two initial ways to mitigate the dataset bias problem in \gls*{lpr}.
The first is leveraging deep learning-based methods’ high capability to visualize and understand how bias has crept into the chosen datasets.
One technique that immediately comes to mind is \gradcam~\cite{selvaraju2017gradcam}, which uses the gradients of any target class flowing into the final convolutional layer to produce a coarse localization map highlighting the important regions in the image for predicting the~class.

The other way is to embrace the ``wildness'' of the internet to collect a large-scale dataset for \gls*{lpr}.
However, as shown in \cref{sec:experiments} and in~\cite{torralba2011unbiased}, downloading images from the internet alone does not guarantee a bias-free sampling, as keyword-based searches return only particular types of images; users of a specific website prefer images with certain characteristics, among other factors.
Thus, such a dataset should be obtained from multiple sources on the internet (e.g., multiple search engines and websites from various~countries).

\balance
\section{Conclusions and Future Work}
\label{sec:conclusions}

In this work, we situate the dataset bias problem~\cite{torralba2011unbiased,tommasi2017deeper} in the \gls*{lpr} context.
We performed experiments on \gls*{lp} images from eight publicly available datasets; four were collected in Brazil and four in mainland China.
The results showed that each dataset does have a unique, identifiable~signature.

Specifically, our \emph{Name That Dataset!} experiments showed that the source dataset of an \gls*{lp} image could be predicted with more than $\bothAccMore$\% accuracy (chance is $1/4$~=~$25$\%).
Intriguingly, we observed no evidence of saturation as more training data was added.
We believe there is no theoretical reason for such results other than the strong biases in the actual~datasets.

Considering our findings, we hope this work will encourage evaluating \gls*{lpr} models in cross-dataset setups, as they provide a better indication of generalization (hence real-world performance) than within-dataset~ones.

How to best leverage the high capability of deep learning-based methods to mitigate the dataset bias problem in \gls*{lpr} is open for research in future work.
As the first step in this direction, we plan to exploit visualization methods, such as \gradcam~\cite{selvaraju2017gradcam}, to produce visually explainable heatmaps to give us clues into why such a lightweight classifier performs so well at distinguishing between \gls*{lp} images from different~datasets.

\ifextended
We also intend to build a large-scale \gls*{lpr} dataset by collecting thousands of images from various internet sources and labeling them with coarse annotations (done automatically).
Although such annotations proved effective for various tasks in other research areas~\cite{liu2020boosting,zheng2020weakly}, work is still lacking on how to best couple coarse-annotated data with fine-annotated data in the \gls*{lpr} context.
We believe that such a dataset could somewhat reduce dataset bias and be employed for training generative models (e.g., GANs) to create an even greater number of \gls*{lp} images from multiple geographical~regions.
\else
\fi

\iffinal
\section*{\uppercase{Acknowledgments}}

\iffinal
    This work was partly supported by the Coordination for the Improvement of Higher Education Personnel~(CAPES) (\textit{Programa de Coopera\c{c}\~{a}o Acad\^{e}mica em Seguran\c{c}a P\'{u}blica e Ci\^{e}ncias Forenses \#~88881.516265/2020-01}), and partly by the National Council for Scientific and Technological Development~(CNPq) (\#~308879/2020-1).
    We gratefully acknowledge the support of NVIDIA Corporation with the donation of the Quadro RTX $8000$ GPU used for this research.
\else
    The acknowledgments are hidden for review.
\fi
\else
\fi

\bibliographystyle{IEEEtran}

\ifextended
\bibliography{bibtex}
\else
\bibliography{bibtex-short}
\fi

\end{document}